\documentclass[twoside,11pt]{article}

\usepackage{jmlr2e}
\usepackage[dvipsnames]{xcolor}
\usepackage{graphicx}
\usepackage{enumerate}
\usepackage{multirow}
\usepackage{listings}
\usepackage{booktabs}
\usepackage{lipsum}
\usepackage{dirtytalk}
\usepackage{pythonhighlight}
\usepackage[flushleft]{threeparttable}

\usepackage{amssymb}

\usepackage{xspace}
\usepackage{pifont}
\providecommand{\cmark}{\textcolor{teal}{\ding{51}}}
\providecommand{\xmark}{\textcolor{red}{\ding{55}}}

\usepackage{mathtools}
\usepackage{amsmath}

\usepackage{tabularx}
\usepackage{multirow}
\usepackage{booktabs}

\newsavebox{\LstBox}

\usepackage{todonotes}

\usepackage{arydshln}

\usepackage{comment}

\jmlrheading{1}{2023}{1-48}{4/00}{10/00}{HierarchicalForecast: A Reference Framework for Hierarchical Forecasting in Python}

\ShortHeadings{HierarchicalForecast: A Reference Framework for Hierarchical Forecasting in Python}{HierarchicalForecast: A Reference Framework for Hierarchical Forecasting in Python}
\firstpageno{1}

\newcommand{\model}[1]{\texttt{#1}}
\newcommand{\HierarchicalForecast}{\model{HierarchicalForecast}}

\newcommand{\BottomUp}{\model{BottomUp}}
\newcommand{\TopDown}{\model{TopDown}}
\newcommand{\MiddleOut}{\model{MiddleOut}}

\newcommand{\BOOTSTRAP}{\model{BOOTSTRAP}}
\newcommand{\NORMALITY}{\model{NORMALITY}}
\newcommand{\PERMBU}{\model{PERMBU}}

\newcommand{\MinTrace}{\model{MinTrace}}
\newcommand{\Comb}{\model{Comb}}
\newcommand{\ERM}{\model{ERM}}

\newcommand{\ARIMA}{\model{ARIMA}}
\newcommand{\ETS}{\model{ETS}}
\newcommand{\GARCH}{\model{GARCH}}

\newcommand{\NumPy}{\model{NumPy}}
\newcommand{\NumBa}{\model{NumBa}}
\newcommand{\ScikitLearn}{\model{sklearn}}

\newcommand{\Pandas}{\model{Pandas}}

\newcommand{\SKTime}{\model{sktime}}
\newcommand{\pmdarima}{\model{pmdarima}}
\newcommand{\DARTS}{\model{darts}}
\newcommand{\TSLearn}{\model{tslearn}}

\newcommand{\Kats}{\model{kats}}

\newcommand{\PyFlux}{\model{pyflux}}
\newcommand{\SegLearn}{\model{seglearn}}

\newcommand{\StatsModels}{\model{statsmodels}}
\newcommand{\StatsForecast}{\model{statsforecast}}

\newcommand{\GluonTS}{\model{gluonts}}
\newcommand{\TFSTS}{\model{tf\_sts}}

\newcommand{\hierarchicalforecast}{\model{hierarchicalforecast}}
\newcommand{\Fable}{\model{fable}}

\newcommand{\HTS}{\model{hts}}
\newcommand{\PyHTS}{\model{pyhts}}
\newcommand{\SKHTS}{\model{scikit-hts}}

\newcommand{\dataset}[1]{\texttt{#1}}
\newcommand{\TourismS}{\dataset{Tourism-S}}
\newcommand{\TourismL}{\dataset{Tourism-L}}
\newcommand{\Labour}{\dataset{Labour}}
\newcommand{\Traffic}{\dataset{Traffic}}
\newcommand{\Wikitwo}{\dataset{Wiki2}}

\long\def\EDIT#1{{\color{black}{#1}\color{black}}}
\begin{document}

\title{HierarchicalForecast: A Reference Framework for Hierarchical Forecasting}

\author{\name Kin G. Olivares* \email kdgutier@cs.cmu.edu \\
        \name Azul Garza* \email federico@nixtla.io \\
        \name David Luo \email djluo@cs.cmu.edu \\
        \name Cristian Challu \email cchallu@cs.cmu.edu \\
        \name Max Mergenthaler-Canseco \email max@nixtla.io \\
        %\name Cristian Challú \email cchallu@cs.cmu.edu \\
        \name Souhaib Ben Taieb \email souhaib.bentaieb@umons.ac.be \\
        \name Shanika L. Wickramasuriya \email s.wickramasuriya@auckland.ac.nz  \\
        \name Artur Dubrawski \email awd@cs.cmu.edu %\\
        }

%\editor{Placeholder Editor}

\maketitle
\def\thefootnote{*}
\footnotetext{These authors contributed equally. 
Corresponding author email address: kdgutier@cs.cmu.edu
}\def\thefootnote{\arabic{footnote}}

\begin{abstract} % <- trailing '%' for backward compatibility of .sty file
Large collections of time series data are commonly organized into structures with different levels of aggregation; examples include product and geographical groupings. It is often important to ensure that the forecasts are coherent so that the predicted values at disaggregate levels add up to the aggregate forecast. The growing interest of the Machine Learning community in hierarchical forecasting systems indicates that we are in a propitious moment to ensure that scientific endeavors are grounded on sound baselines. For this reason, we put forward the \HierarchicalForecast\ library, which contains preprocessed publicly available datasets, evaluation metrics, and a compiled set of statistical baseline models. Our Python-based reference framework aims to bridge the gap between statistical and econometric modeling, and Machine Learning forecasting research.
%A necessary condition for \say{coherent} decision-making and planning, with such datasets, is for the dis-aggregated series' forecasts to add up exactly to the aggregated series forecasts, which motivates the creation of novel hierarchical forecasting algorithms.

\end{abstract}


%%%%%%%%%%%
\section{Introduction} \label{section:introduction}

Multivariate time series data can often be organized into hierarchical structures with different levels of aggregation. Independently forecasting all the series is unlikely to produce \emph{coherent} forecasts, that is, forecasts which satisfy the aggregation constraints as the original data. In a nutshell, \href{https://nixtla.github.io/hierarchicalforecast/examples/introduction.html}{\textcolor{BlueViolet}{hierarchical time series forecasting}} is a multitask forecasting problem with a set of linear aggregation constraints to be satisfied.

While summing forecasts for the most disaggregated level (called \emph{bottom-up}) will provide coherent forecasts, it can perform poorly on highly dissagregated series. Novel hierarchical forecasting methods first generate independent forecasts for each series (called \emph{base} forecasts), then reconcile them to produce coherent forecasts \citep{wickramasuriya2019hierarchical_mint_reconciliation}.

There is substantial interest on Hierarchical Forecasting from both industry and academia, as shown by the international forecasting competitions GEFCOM2012 \citep{hong2014gefcom} and M5 \citep{makridakis2021m5competition}, and the Machine Learning (ML) community's growing interest in the topic \citep{rangapuram2021hierarchical_e2e, xing2021sharq_hierarchical,paria2021hierarchical_hired,olivares2021neural_hierarchical_forecasting,kamarthi2022profhit_network,panagiotelis2023probabilistic_reconciliation}. 
%\KO{Who else do we cite?}

An enabling condition for the systematic development of useful forecasting methods is the ability to empirically evaluate and compare newly proposed methods with state-of-the-art and well-established baselines. However, ML research on hierarchical forecasting faces two challenges. First, while Python continues to grow in popularity among the ML community \citep{piatetsky2018python_survey}, it lacks many relevant statistical and econometric \EDIT{modeling} packages (often originally developed in the R language). As a result, researchers must build Python bridges to access the R baselines' implementations. Rapid, substantial development in statistical methods for hierarchical forecasting exacerbates this problem. Secondly, the Python global interpreter lock limits its programs to use a single thread, which prevents us from taking advantage of the available multi-core resources to speed up the software. When implemented naively, statistical baselines in Python take excessively long execution times, surpassing those of more complex methods, and discouraging their use.

We introduce the open-source benchmark library \HierarchicalForecast to tackle these challenges\footnote{License: CC-by 4.0, see \href{https://creativecommons.org/licenses/by/4.0/}{\textcolor{BlueViolet}{https://creativecommons.org/licenses/by/4.0/}}.\\
Code and documentation are available in  \href{https://github.com/Nixtla/hierarchicalforecast}{\textcolor{BlueViolet}{https://github.com/Nixtla/hierarchicalforecast}}.}. Our work builds upon Python's fastest open-source  
\ETS/\ARIMA\footnote{Autoregressive Integrated Moving Average (\ARIMA) and Exponential Smoothing (\ETS) are two of the most important univariate forecasting baseline methods.}  implementations and well-performing neural forecasting methods to improve the availability, utility, and adoption of hierarchical forecast reference baselines.

%%%%%%%%%%%
\section{Library description (/features)}

Compared to existing hierarchical forecasting software libraries, \HierarchicalForecast\ has the following distinctive features:

\textbf{Minimal dependencies}. Our library is built with minimal dependencies using \NumPy\ for linear algebra and array operations \citep{harris2020numpy}, \Pandas\ for data manipulation \citep{mckinney2010pandas} and \ScikitLearn\ for predictive modeling \citep{scikit-learn}. We compute base forecasts using the \StatsForecast\ package~\citep{garza2022statsforecast}, which provides the fastest implementations of Auto\ARIMA\ and Auto\ETS\ \citep{hyndman2008automatic_arima} based on \NumBa\ \citep{lam2015numba}. This just-in-time compiler optimizes Python's \NumPy\ code to reach execution speed attainable with native C language code.

\textbf{Comprehensive set of hierarchical forecasting methods}. Some hierarchical forecasting Python implementations are available in the following packages: \GluonTS~\citep{alexandrov2020package_gluonts}, \DARTS~\citep{herzen2022darts_timeseries}, \SKHTS~\citep{mazzaferro2022skhts_package}, \SKTime~\citep{loning2019sktime}, and \PyHTS~\citep{zhang2022hts_package}. However, as seen in Table~\ref{table:software_availability}, each of these libraries only hosts a subset of the State-Of-The-Art (SOTA) methods. Our library provides unified access to a comprehensive set of these methods and enables robust performance validation of the implementations to ensure the Python community's access to efficient and reliable baselines. \HierarchicalForecast's \href{https://nixtla.github.io/hierarchicalforecast/methods.html}{\textcolor{BlueViolet}{curated collection}} of reference algorithms includes \BottomUp~\citep{orcutt1968hierarchical_bottom_up, dunn1976hierarchical_bottom_up2}, \TopDown~\citep{gross1990hierarchical_top_down, fliedner1999hierarchical_top_down2}, \MiddleOut, \MinTrace~\citep{wickramasuriya2019hierarchical_mint_reconciliation}, and \ERM~\citep{taieb2019hierarchical_regularized_regression}\ for point forecasting, and it is the only Python library so far that includes SOTA probabilistic forecasting methods, including \PERMBU~\citep{taieb2017coherent_prob_forecasts}, \NORMALITY~\citep{wickramasuriya2023probabilistic_gaussian}, and \BOOTSTRAP~\citep{panagiotelis2023probabilistic_reconciliation}.
% Being open, it allows straightforward expansion of the current set of methods by its users.
%\MinTrace\footnote{We improve \MinTrace's efficiency with Equation (9) solution in \citet{wickramasuriya2019optimal}.}

% \newpage
\textbf{Forecast evaluation and visualization}. Our library facilitates a complete forecast evaluation across the levels of the hierarchical structure. It includes multiple standard accuracy measures for point forecasts. Furthermore, it also includes multiple scoring rules to evaluate probabilistic forecasts, such as \EDIT{the multivariate logarithmic and energy scores ~\citep{panagiotelis2023probabilistic_reconciliation} and the univariate scaled continuous ranked probability score (sCRPS)~\citep{gneiting2011crps,olivares2021neural_hierarchical_forecasting,makridakis2022m5_uncertainty}}. In addition to the forecast accuracy \href{https://nixtla.github.io/hierarchicalforecast/evaluation.html}{\textcolor{BlueViolet}{evaluation tools}}, the package provides specialized \href{https://nixtla.github.io/hierarchicalforecast/utils.html}{\textcolor{BlueViolet}{visualization tools}}.

\textbf{Hierarchical time series datasets}. The library \EDIT{provides access to five \Pandas\ datasets and the aggregation utils to create them}. Australian \Labour\ monthly reports~\citep{Aulabor2019Aulabor_dataset}, SF Bay Area daily \Traffic\ measurements~\citep{dua2017traffic_dataset},
Quarterly Australian \TourismS\ visits~\citep{canberra2005tourismS_dataset}, Monthly Australian \TourismL\ visits~\citep{canberra2019tourismL_dataset}, and daily \Wikitwo\ article views~\citep{wikipedia2018web_traffic_dataset}. Each dataset is accompanied by metadata capturing its seasonality/frequency, the forecast horizon used in previous publications, its corresponding hierarchical aggregation constraints matrix, and the names of its levels.
% Multiple hierarchical forecasting studies have used these datasets in the past \citep{wickramasuriya2019hierarchical_mint_reconciliation, taieb2019hierarchical_regularized_regression, rangapuram2021hierarchical_e2e, olivares2021neural_hierarchical_forecasting}.

% \begin{table*}[htp]
\begin{table*}[t]
\tiny
%\scriptsize
%\footnotesize
\centering
\caption{Availability of hierarchical forecasting methods in considered software libraries.} \label{table:software_availability}
{
% \color{blue}
\setlength\tabcolsep{2.8pt}
\begin{tabular}{cl|cccccccccccc|ccc|}
\toprule
	&       & \multicolumn{12}{c|}{Point F. Methods}                                                                                                                                                                                                                         &  \multicolumn{3}{c|}{Probabilistic F. Methods}   \\ %\cline{3-17}
	&                       & \multicolumn{1}{c|}{\BottomUp} & \multicolumn{3}{c|}{\TopDown}  & \multicolumn{1}{c|}{\MiddleOut} & \multicolumn{2}{c|}{\Comb}  & \multicolumn{3}{c|}{\MinTrace}                                   & \multicolumn{2}{c}{\ERM}                  & \BOOTSTRAP & \NORMALITY & \PERMBU                \\
	& \textbf{Library}		& \multicolumn{1}{c|}{}          & (f)         & (a)              & \multicolumn{1}{c|}{(p)}        &  \multicolumn{1}{c|}{(f)}   & (ols) & \multicolumn{1}{c|}{(wls)} & (ols) & (ws)  & \multicolumn{1}{c|}{(wv)} & (cf)         & (lasso)      &            &            &                        \\ \midrule
\parbox[t]{.2mm}{\multirow{4}{*}{\rotatebox[origin=c]{90}{Python}}}
	& \hierarchicalforecast &\cmark                          &\cmark       &\cmark            &\cmark         &\cmark           &\cmark                       &\cmark                              &\cmark &\cmark &\cmark                     &\cmark        &\cmark        &\cmark      &\cmark      &\cmark                  \\
	& \DARTS/hts            &\cmark                          &\cmark       &\xmark            &\xmark         &\xmark           &\cmark                       &\cmark                              &\cmark &\cmark &\cmark                     &\xmark        &\xmark        &\xmark      &\xmark      &\xmark                  \\
	& \SKHTS                &\cmark                          &\xmark       &\xmark            &\xmark         &\xmark           &\cmark                       &\cmark                              &\cmark &\cmark &\cmark                     &\xmark        &\xmark        &\xmark      &\xmark      &\xmark                  \\
	& \PyHTS                &\cmark                          &\xmark       &\xmark            &\xmark         &\xmark           &\cmark                       &\cmark                              &\cmark &\cmark &\cmark                     &\xmark        &\xmark        &\xmark      &\xmark      &\xmark                  \\ \midrule
\parbox[t]{.2mm}{\multirow{3}{*}{\rotatebox[origin=c]{90}{R}}}
	& \Fable                &\cmark                          &\cmark       &\cmark            &\cmark         &\cmark           &\cmark                       &\cmark                              &\cmark &\cmark &\cmark                     &\xmark        &\xmark        &\xmark      &\xmark      &\xmark                  \\
	& \HTS                  &\cmark                          &\cmark       &\cmark            &\cmark         &\cmark           &\cmark                       &\cmark                              &\cmark &\cmark &\cmark                     &\xmark        &\xmark        &\cmark      &\cmark      &\xmark                  \\
	& \GluonTS/\HTS         &\cmark                          &\xmark       &\xmark            &\xmark         &\xmark           &\cmark                       &\cmark                              &\cmark &\cmark &\cmark                     &\cmark        &\cmark        &\xmark      &\xmark      &\cmark                  \\
\bottomrule
\end{tabular}
}
\end{table*}

\section{Related Software}

We refer to~\cite{januschowski2019opensource_forecasting}, and~\cite{siebert2021forecasting_survey} for complete open-source forecasting software surveys. We describe packages relevant to \HierarchicalForecast. Classic statistical and econometric time series models, such as \ARIMA, \ETS, and \GARCH\ have low-level \NumPy\ implementations in multiple libraries, including \StatsModels, \TFSTS, \Kats, and \PyFlux. Currently, the \StatsForecast\ package provides the fastest \NumBa\ implementations of these methods, and it has been adopted in multiple popular open-source Python frameworks, such as \DARTS\ and \SKTime. Other libraries, including \GluonTS\ have Python API R-connections that enable comparisons with well-established methods at the cost of R dependency frictions. Finally, higher-level libraries, including \DARTS, \SKTime, \TSLearn, \pmdarima, and \SegLearn, provide API access to various time series forecasting models. %, \TFSTS, \FlowForecast, and \NeuralForecast,

% \clearpage
\section{Usage Example and Benchmarks}

The code example below highlights the usability and wide rate of available reconciliation methods in \HierarchicalForecast. It predicts eight months of the 57 series of the \Labour\ dataset using Auto\ARIMA\ base model and later \EDIT{reconciles} the base predictions using the \BottomUp, \TopDown, and \MinTrace\ methods. We generate prediction intervals with 80\% and 90\% coverage using the \BOOTSTRAP\ technique.

% \captionsetup[table]{skip=0pt}
% \begin{table*}[htp]
\begin{table*}[t]
\tiny
\centering
\setlength\tabcolsep{2.3pt}
\caption[short caption]{
Mean sCRPS of \ARIMA-based coherent forecasts, 95\% confidence, 10 seeds. 
% Mean sCRPS, for 10 random seeds, with 95\% confidence intervals.
% \EDIT{Deprecate BU, in favor for new three columns in "Other" category base, mlforecast, nbeats
% \\
% base ARIMA \\
% recNBEATS = NBEATS + MinTRace + Bootstrap \\
% recXGbooost = XGbooost + MinTRace + Bootstrap
% }
}
\label{table:crps_evaluation}
\begin{threeparttable}
\begin{tabular}{l | ccc | ccc | ccc}
\toprule
                  &              &    \BOOTSTRAP &              &                     &\NORMALITY       &                     &              &    \PERMBU\tnote{1}   &               \\
 \textsc{Dataset} &    \EDIT{\BottomUp} &     \TopDown\tnote{1,2}  &    \MinTrace &     \EDIT{\BottomUp}       &\TopDown\tnote{1,2}         &    \MinTrace        &    \EDIT{\BottomUp} &    \TopDown\tnote{1}  &     \MinTrace \\ \midrule 
          \Labour &  0.006±0.001 &  0.067±0.001  &  0.006±0.001 &  0.007±0.000        &  0.067±0.000    &  0.006±0.000        &  0.006±0.001 &  0.063±0.001 &  0.006±0.001  \\ 
         \Traffic &  0.067±0.002 &  0.070±0.001  &  0.050±0.001 &  0.076±0.000        &  0.071±0.000    &  0.055±0.000        &  0.076±0.001 &  0.063±0.001 &  0.052±0.001  \\ 
        \TourismS &  0.089±0.001 &  0.120±0.001  &  0.089±0.001 &  0.083±0.000        &  0.117±0.000    &  0.086±0.000        &  0.084±0.001 &  0.103±0.001 &  0.084±0.001  \\ 
        \TourismL &  0.142±0.001 &          -    &  0.131±0.001 &  0.169±0.000        &    -            &  0.133±0.000        &          -   &          -   &          -    \\ 
         \Wikitwo &  0.417±0.002 &  0.356±0.005  &  0.352±0.002 &  0.509±0.000        &  0.328±0.000    &  0.398±0.000        &  0.512±0.003 &  0.423±0.005 &  0.459±0.003  \\ \bottomrule
\end{tabular}
\end{threeparttable}
\begin{tablenotes}
    \item[1] \textsuperscript{1} \TopDown/\PERMBU\ results are unavailable because, they cannot be applied to group hierarchical structures.
    \item[2] \textsuperscript{2} The combinations \NORMALITY-\TopDown\ and \BOOTSTRAP-\TopDown\ are yet to be implemented, this has never been done before.
\end{tablenotes}
\end{table*}
% \vspace{0.00mm}

\vspace{40pt}

\begin{python}
from statsforecast.core import StatsForecast
from statsforecast.models import AutoARIMA
from datasetsforecast.hierarchical import HierarchicalData
from hierarchicalforecast.core import HierarchicalReconciliation
from hierarchicalforecast.evaluation import HierarchicalEvaluation
from hierarchicalforecast.methods import BottomUp, TopDown, MinTrace

# Load Labour dataset
Y_df, S_df, tags = HierarchicalData.load('./data', 'Labour')
Y_df = Y_df.set_index('unique_id')
# Compute base AutoARIMA predictions and reconcile them
fcst = StatsForecast(df=Y_df, models=[AutoARIMA(season_length=12)], 
                     freq='MS', n_jobs=-1)
Y_hat_df = fcst.forecast(h=8, fitted=True)
Y_fitted_df = fcst.forecast_fitted_values()
# Define reconcilers
reconcilers = [BottomUp(), 
               TopDown(method='average_proportions'), 
               MinTrace(method='ols')]
# Reconcile
hrec = HierarchicalReconciliation(reconcilers=reconcilers)
Y_rec_df = hrec.reconcile(Y_hat_df, S_df, tags, Y_df=Y_fitted_df,
                          intervals_method='bootstrap', level=[80,90])
\end{python}

Table~\ref{table:crps_evaluation} shows the overall average sCRPS for various \HierarchicalForecast\ reconciliation methods, along with the measurement's 95\% confidence intervals for the five datasets. These \href{https://github.com/Nixtla/hierarchicalforecast/blob/main/nbs/examples/TourismLarge-Evaluation.ipynb}{\textcolor{BlueViolet}{experimental results}} are aligned with previous studies' reports~\citep{wickramasuriya2019hierarchical_mint_reconciliation, taieb2019hierarchical_regularized_regression,rangapuram2021hierarchical_e2e, olivares2021neural_hierarchical_forecasting}.

\section{Conclusion and Plans}

We present \HierarchicalForecast, a Python open-source library dedicated to hierarchical time series forecasting. The library integrates publicly available processed datasets, evaluation metrics, and a curated set of highly efficient statistical baselines. We provide examples and references to extensive experiments to show how to use the baselines and evaluate their empirical performance. This work will help the Machine Learning forecasting community by bridging the gap between statistical and econometric modeling and providing benchmark tools for developing novel hierarchical forecasting algorithms compared to the well-established methods. We intend to continue maintaining and improving the repository and promoting collaboration across the forecasting research community.
% \KO{Hint to future work on Neural + Hierarchical, Distributed computation}

\clearpage
\section*{Acknowledments} \label{section:acknowledgements}

This work was partially supported by the Defense Advanced Research Projects Agency (award FA8750-17-2-0130), the National Science Foundation (grant 2038612), the Space Technology Research Institutes grant from NASA’s Space Technology Research Grants Program, the U.S. Department of Homeland Security (award 18DN-ARI-00031), and by the U.S. Army Contracting Command (contracts W911NF20D0002 and W911NF22F0014 delivery order \#4). The Fonds de la Recherche Scientifique supported this work – FNRS under Grant No J.0011.20. Thanks to Pedro Mercado, Syama Rangapuram, and Chirag Nagpal for the in-depth discussion and comments on the literature and library. The authors also thank Shibo Zhou and José Morales for their software contributions.

% \clearpage
\bibliography{citations}  
\end{document}